%% file: main.tex
\documentclass[10pt,twocolumn,letterpaper]{article}

\usepackage{cvpr}              
\usepackage{pifont}
\usepackage[table]{xcolor}
\usepackage{caption} 
\usepackage{setspace}
\usepackage{amsmath} 

\definecolor{cvprblue}{rgb}{0.21,0.49,0.74}
\usepackage[pagebackref,breaklinks,colorlinks,allcolors=cvprblue]{hyperref}

\def\paperID{xxx} 
\def\confName{CVPR}
\def\confYear{2026}

\title{Dr.Occ: Depth- and Region-Guided 3D Occupancy from Surround-View Cameras for Autonomous Driving}

\author{
    Xubo Zhu$^{1,2,*}$, Haoyang Zhang$^{2,\#}$, Fei He$^2$, Rui Wu$^2$, Yanhu Shan$^2$, Wen Yang$^1$, Huai Yu$^{1,3,\dag}$ \\
    $^1$School of Electronic Information, Wuhan University \quad $^2$Horizon Robotics \\
    \quad $^3$Wuhan University Shenzhen Research Institute \\
    {\tt \{zhuxubo, yangwen, yuhuai\}@whu.edu.cn} \\
    {\tt \{haoyang.zhang, fei.he, rui.wu, yanhu.shan\}@horizon.ai}
}

\definecolor{Gray}{gray}{0.95}
\definecolor{barriercolor}{RGB}{119, 136, 153}
\definecolor{bicyclecolor}{RGB}{139, 166, 193}
\definecolor{buscolor}{RGB}{175, 198, 203}
\definecolor{carcolor}{RGB}{100, 149, 236}
\definecolor{constructcolor}{RGB}{145, 170, 157}
\definecolor{motorcolor}{RGB}{186, 140, 172}
\definecolor{pedestriancolor}{RGB}{72, 61, 139}
\definecolor{trafficcolor}{RGB}{218, 165, 33}
\definecolor{trailercolor}{RGB}{205, 133, 63}
\definecolor{truckcolor}{RGB}{205, 133, 63}
\definecolor{driveablecolor}{RGB}{169, 169, 169}
\definecolor{otherflatcolor}{RGB}{205, 192, 176}
\definecolor{sidewalkcolor}{RGB}{191, 184, 176}
\definecolor{terraincolor}{RGB}{143, 188, 143}
\definecolor{manmadecolor}{RGB}{222, 184, 135}
\definecolor{vegetationcolor}{RGB}{85, 107, 47}
\definecolor{otherscolor}{RGB}{0, 0, 0}

\makeatletter  
\renewcommand{\@makefntext}[1]{%
  \noindent\makebox[0pt][r]{}#1  
}
\makeatother 

\begin{document}
\twocolumn[{ %
\renewcommand\twocolumn[1][]{#1}%
\maketitle
\begin{center} \vspace{-2.2em}
  \centering
  \includegraphics[width=1.0\linewidth]{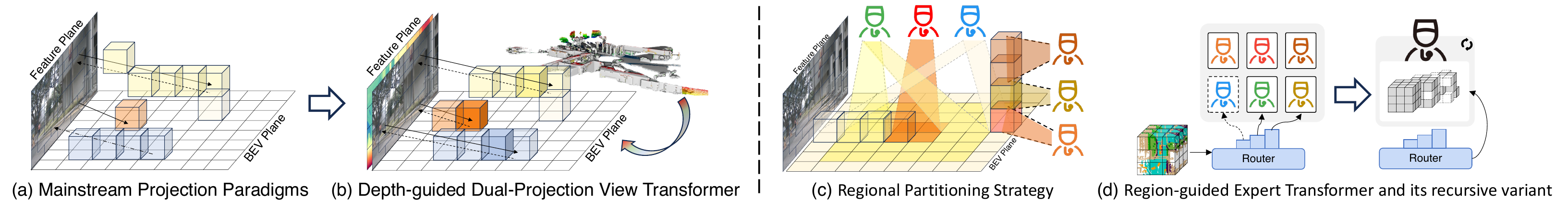}
  \vspace{-2.0em}
  \captionof{figure}{\setstretch{0.9}In \textbf{(a)}, we illustrate mainstream projection paradigms in vision-based 3D perception. In \textbf{(b)}, we propose a dual-projection scheme that leverages high-quality fine-grained depth cues to enhance geometric feature representation. In \textbf{(c-d)}, we further design an MoE/MoR-style Transformer, adaptively assigning region-specific experts to capture different spatial regions based on distance and height.}
  \label{fig:teaser}
\end{center}
}]
\renewcommand{\thefootnote}{}
\footnotetext{\noindent $^*$Work done during internship at Horizon Robotics. $^\#$Project Lead. \\$\dag$ Corresponding author.}
\input{sec/0_abstract}
\input{sec/1_intro}

\input{sec/2_relatedworks}

\input{sec/3_method}

\input{sec/4_experiments}

\input{sec/5_conclusion}
{
    \small
    \bibliographystyle{ieeenat_fullname}
    \bibliography{main}
}

\end{document}

%% file: sec/0_abstract.tex
\begin{abstract}
3D semantic occupancy prediction is crucial for autonomous driving perception, offering comprehensive geometric scene understanding and semantic recognition. However, existing methods struggle with geometric misalignment in view transformation due to the lack of pixel-level accurate depth estimation, and severe spatial class imbalance where semantic categories exhibit strong spatial anisotropy. To address these challenges, we propose \textbf{Dr.Occ}, a depth- and region-guided occupancy prediction framework. Specifically, we introduce a depth-guided 2D-to-3D View Transformer (D$^2$-VFormer) that effectively leverages high-quality dense depth cues from MoGe-2 to construct reliable geometric priors, thereby enabling precise geometric alignment of voxel features. Moreover, inspired by the Mixture-of-Experts (MoE) framework, we propose a region-guided Expert Transformer (R/R$^2$-EFormer) that adaptively allocates region-specific experts to focus on different spatial regions, effectively addressing spatial semantic variations.
Thus, the two components make complementary contributions: depth guidance ensures geometric alignment, while region experts enhance semantic learning.
Experiments on the Occ3D--nuScenes benchmark demonstrate that \textbf{Dr.Occ} improves the strong baseline BEVDet4D by 7.43\% mIoU and 3.09\% IoU under the full vision-only setting.

\end{abstract}

%% file: sec/1_intro.tex
\section{Introduction}
\label{sec:intro}
3D semantic \textbf{OCC}upancy (OCC) prediction has emerged as a cornerstone task in autonomous driving and robotics, aiming to produce dense and metrically accurate volumetric maps of surrounding environments ~\cite{huang2023tri, tian2023occ3d, liu2023fully, ma2024cotr}. 
This fine-grained voxel-wise occupancy representation offers rich geometric and semantic cues for safe navigation, motion planning, and collision avoidance in complex scenarios~\cite{tesla_ai_day}. However, achieving geometrically and semantically accurate occupancy estimation demands holistic reasoning over visible, occluded, and dominant free-space regions, posing significant challenges to current methods, especially for vision-based approaches~\cite{li2023fbocc, xu2025survey}.

In particular, vision-based OCC methods face two main challenges: reconstructing accurate 3D geometry from 2D observations, and learning reliable semantics from heavily imbalanced volumetric labels. As illustrated in~\cref{fig:teaser}(a-c), existing 2D-to-3D view transformer methods typically address the geometric challenge through BEV-based intermediate representations, employing various projection strategies: forward-projection approaches (LSS~\cite{philion2020lift}, BEVDepth~\cite{li2023bevdepth}, BEVStereo~\cite{li2023bevstereo}) that predict depth at low resolution to lift 2D features into 3D space, backward-projection frameworks (BEVFormer~\cite{li2022bevformer}) that leverage transformer-based warping to map BEV plane back to image views, and dual-projection schemes (FBOCC~\cite{li2023fbocc}, COTR~\cite{ma2024cotr}) that integrate both directions by densifying forward-projection features through backward-projection. 
However, these methods primarily rely on low-resolution, noisy depth estimates for 2D-to-3D feature transformation, which limits the accuracy of feature mapping. 

In contrast, with the rapid development of large vision models, more advanced 3D geometry estimation models are now publicly available~\cite{wang2025moge, wang2025moge2}. We believe that leveraging these advanced depth estimation models will greatly benefit the 3D OCC task, as they offer significantly improved geometry recovery. However, we find that effectively applying these models to the OCC task is not straightforward. Simply concatenating the depth map with the image does not yield satisfactory results, nor does the approach of converting estimated depth into pseudo point clouds.

After extensive experimentation and analysis, we propose an effective approach to leverage the advanced 3D estimation models. Specifically, we introduce a \textbf{Depth-guided 2D-to-3D View Transformer (D$^2$-VFormer)} that utilizes robust geometric priors to generate a geometry-aware occupancy mask, which identifies non-empty voxel locations. This mask serves as a strong inductive bias that guides the model to focus computational resources on semantically meaningful regions, rather than wasting effort on empty space. As a result, the OCC model achieves more precise and effective 2D-to-3D feature mapping.

On the semantic side, vision-based OCC must handle highly imbalanced voxel distributions. To mitigate this imbalance, class-rebalancing strategies such as focal and weighted losses~\cite{lin2017focal} are widely adopted, and some existing approaches~\cite{chen2023group, ma2024cotr} further resort to coarse-to-fine semantic grouping or hierarchical semantic strategies. 

To delve deeper into this imbalance problem, we observe that different semantic classes exhibit strong positional preferences in 3D space. For example, pedestrians tend to concentrate near road boundaries, vehicles cluster in road centers, while buildings and vegetation are found at higher elevations. This spatial anisotropy of semantics suggests that model capacity should be allocated adaptively across regions, rather than uniformly. Leveraging this observation and inspired by the Mixture-of-Experts (MoE) framework~\cite{zhou2022moe}, we propose a \textbf{Region-guided Expert Transformer (R-EFormer)} and its recursive variant, \textbf{R$^2$-EFormer}, which adaptively allocates region-specific experts to focus on different spatial regions, effectively addressing spatial semantic variations.

Building upon these innovations, we propose \textbf{Dr.Occ}, a unified vision-based 3D occupancy prediction framework that tackles both geometric and semantic challenges through depth-guided feature mapping and region-guided semantic modeling, achieving geometrically consistent and semantically balanced occupancy estimation.

Extensive experiments on the Occ3D benchmark~\cite{tian2023occ3d} show that Dr.Occ significantly improves various baseline models. 
When integrated into the strong baseline BEVDet4D~\cite{huang2022bevdet4d}, Dr.Occ achieves substantial improvements of 7.43\% mIoU and 3.09\% IoU. Furthermore, when our modules are integrated into the SOTA method COTR~\cite{ma2024cotr}, they further boost the overall mIoU by 1.0\%, demonstrating the strong generalizability and effectiveness of our unified design for joint geometric–semantic occupancy prediction.



%% file: sec/2_relatedworks.tex
\section{Related Work}
\label{sec:formatting}
\subsection{2D-to-3D View Transformation}
Transforming 2D image features into 3D representations is fundamental to vision-based occupancy prediction. Existing methods typically follow one of three transformation paradigms. The first is forward projection, which lifts image features along camera rays and aggregates them into a BEV map or a voxel grid. Representative methods include Inverse Perspective Mapping (IPM)~\cite{mallot199ipm}, which projects features onto a predefined plane, and Lift Splat Shoot (LSS)~\cite{philion2020lift, li2023bevdepth, li2023bevstereo}, which explicitly estimates depth to lift perspective-view features into a three-dimensional lattice. Alternatively, backward projection places queries at top-down or volumetric locations and samples evidence by projecting those coordinates into the image views, as explored by BEVFormer~\cite{li2022bevformer}, VoxFormer~\cite{li2023voxformer}, and TPVFormer~\cite{huang2023tri}. The third is a hybrid forward–backward design that couples the two directions to enforce cross-view consistency, exemplified by FBOcc~\cite{li2023fb} and COTR~\cite{ma2024cotr}. 

Despite these advances, current pipelines rely on implicit ray-based geometry estimation at low resolution, which is ill-posed under monocular or limited multi-view settings, inducing projection errors and feature misalignment. In contrast, we leverage high-quality pixel-level depth from pretrained models (e.g., MoGe-2~\cite{wang2025moge2}) to construct precise geometric priors, enabling dense and geometrically accurate occupancy reconstruction.
\begin{figure*}[!t]
    \centering
    \includegraphics[width=0.7\textwidth]{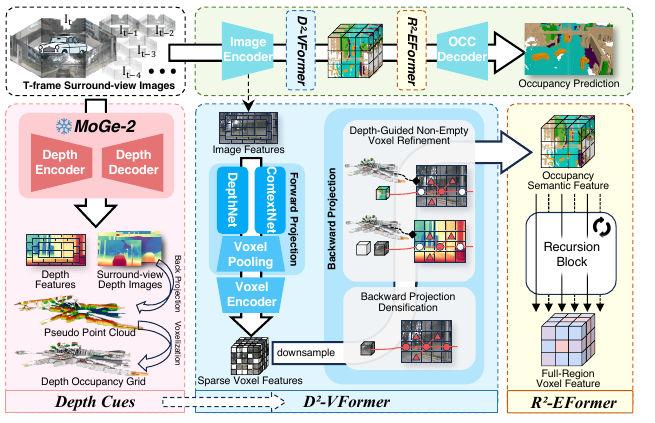}
    \vspace{-1.0em}
    \caption{\setstretch{0.9}\textbf{The overall architecture of Dr.Occ.} T consecutive surround‑view images are processed by MoGe‑2~\cite{wang2025moge2} to estimate depth maps, which provide geometric cues for D\textsuperscript{2}-VFormer to construct dense, low‑cost, and geometrically accurate voxel features. These features are then refined in R\textsuperscript{2}-EFormer through recursive semantic decoding. Finally, the refined features are decoded by the OCC Decoder to produce the occupancy prediction.}
    \label{fig:arch}

\end{figure*}
\subsection{Vision-based 3D Occupancy Prediction}
To address the limited vertical expressivity of BEV maps and improve coverage of general obstacles, recent work advances 3D voxel-level occupancy prediction, which encodes both free space and occupied regions throughout the scene. 3D semantic scene completion~\cite{cao2022monoscene} is the precursor to 3D occupancy prediction, defining the core task of estimating a semantic voxel grid from visual input, and early methods such as MonoScene~\cite{cao2022monoscene}, OccFormer~\cite{occformer}, and VoxFormer~\cite{li2023voxformer} established strong baselines. Subsequent research introduces a self-supervised paradigm that unites 3D Gaussian Splatting with neural radiance fields, as exemplified by SelfOcc~\cite{SelfOcc}, OccNeRF~\cite{OccNeRF}, GaussianOcc~\cite{GaussianOcc}, and GaussTR~\cite{GaussTR}, to learn geometry and semantics from multi-view consistency. Another thread prioritizes efficiency and effectiveness. FlashOcc~\cite{flashocc}, SparseOcc~\cite{liu2023fully}, and COTR~\cite{ma2024cotr} reduce redundancy and improve occupancy quality through sparse reasoning and cross-view constraints. Additionally, a line of work represented by EfficientOCF~\cite{xu2025eff} exploits flow cues, whereas PanoOcc~\cite{wang2024panoocc} incorporates instance-level priors, improving accuracy and robustness. Finally, diffusion-based formulations have been explored for occupancy synthesis and refinement, exemplified by OccGen~\cite{wang2024occgen} and DiffOcc~\cite{wang2025diffocc}. 

However, despite recent progress, the long-tail problem induced by skewed semantic distributions remains largely unaddressed. Existing approaches overlook the spatial clustering characteristics of rare classes, treating all regions uniformly and thus limiting recall for rare categories. In contrast, we leverage region-wise expert decomposition to adaptively model spatial clustering patterns, enabling balanced and fine-grained semantic learning.

\subsection{Mixture of Experts and Mixture of Recursions} 
Mixture-of-Experts (MoE) has been widely adopted for efficient model scaling by dynamically routing inputs to specialized experts~\cite{zhou2022moe}. Recently, Mixture-of-Recursions (MoR)~\cite{bae2025mor} extends this paradigm by replacing multiple experts with a single expert applied recursively, reducing parameters while maintaining representational capacity. Inspired by these principles, we adapt MoE~\cite{zhou2022moe} and MoR~\cite{bae2025mor} concepts to 3D occupancy prediction, designing region-specific experts that progressively refine semantic predictions in spatially coherent regions.

%% file: sec/3_method.tex
\section{Methods}
\label{sec:method}
\subsection{Problem Formulation}
We consider the task of vision-based 3D semantic occupancy prediction, where the system observes $T$ consecutive frames from $N_c$ calibrated surround-view cameras. For each camera $i \in \{1, \dots, N_c\}$, the intrinsic matrix is denoted as $\mathbf{K}_i \in \mathbb{R}^{3\times 3}$ and the extrinsic parameters as $\left[ \mathbf{R}_i|\mathbf{t}_i \right]$, where $\mathbf{R}_i \in \mathbb{R}^{3\times 3}$ is the rotation matrix and $\mathbf{t}_i \in \mathbb{R}^3$ is the translation vector. At each time step $t \in [1, T]$, the multi-camera image set is represented as $\{\mathbf{I}_1^{(t)}, \mathbf{I}_2^{(t)}, \dots, \mathbf{I}_{N_c}^{(t)}\}$, where $\mathbf{I}_i^{(t)} \in \mathbb{R}^{H \times W \times 3}$ denotes an image of spatial resolution $H \times W$ from camera $i$.
The goal is to predict a 3D voxel-wise semantic occupancy grid 
$\mathbf{O} \in \mathbb{R}^{X \times Y \times Z \times C}$ 
describing the environment around the ego-vehicle, 
where $(X,Y,Z)$ denote the spatial resolution and $C$ is the number of semantic classes. 
Following the prevailing paradigm~\cite{flashocc, huang2022bevdet4d, ma2024cotr}, our pipeline first extracts 2D image features 
$\mathbf{F}_{2\mathrm{D}}^{(t)} \in \mathbb{R}^{H \times W \times D}$ 
from each frame $t \in [1, T]$ of the multi-camera input, 
then applies various transformers $\{\mathcal{T}\}$ to lift and aggregate features across all $T$ frames 
into a unified 3D feature volume $\mathbf{F}_{3\mathrm{D}} \in \mathbb{R}^{X \times Y \times Z \times D}$. 
Finally, a 3D decoder $\mathcal{D}$ maps $\mathbf{F}_{3\mathrm{D}}$ 
to the semantic occupancy prediction 
$\hat{\mathbf{O}} = \mathcal{D}(\mathbf{F}_{3\mathrm{D}})$.

\subsection{Overall Architecture}
As illustrated in Fig.~\ref{fig:arch}, the architecture of \textbf{Dr.Occ} follows the general pipeline of prior mainstream approaches while making two key contributions: a redesigned view transformer \textbf{D$^2$-VFormer} $\mathcal{T}_{v}$ that leverages high-quality depth cues, and an MoE-style region-guided semantic refinement module \textbf{R$^2$-EFormer} $\mathcal{T}_{s}$.
Notably, the high-quality depth cues are obtained from MoGe-2~\cite{wang2025moge2}, which provides both dense depth features and pixel-level depth estimates.

\subsection{Image Encoder}
At time step $t$, for the image $\mathbf{I}_i^{(t)}$ captured by camera $i$, 
a ResNet-50~\cite{resnet} backbone is used to extract its feature map $\mathbf{F}_i^{(t)}$. 
Collecting features from all $N_c$ cameras over $T$ frames yields the feature set 
$\{\mathbf{F}_i^{(t)} \mid i \in [1, N_c], \ t \in [1, T]\}$, 
which can be divided into features from the current frame and those from past frames. 
These features are subsequently utilized in the following modules for forward projection to generate voxel features. Notably, the current-frame feature set ${F}^{(I)} = \{\mathbf{F}_i^{(T)} \mid i \in [1, N_c]\}$ is additionally preserved and fed to subsequent modules as semantic cues to provide fine-grained image-level guidance for semantic refinement.

\subsection{Depth-guided Geometric Enhancement}
\begin{figure}[t]
    \centering
    \includegraphics[width=0.85\linewidth]{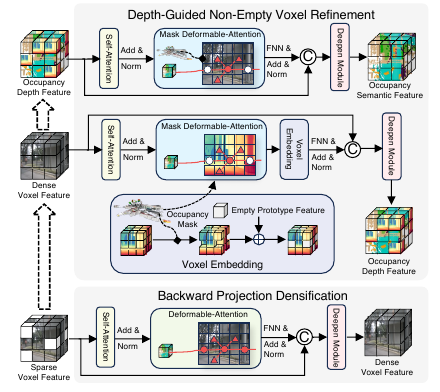}
    \caption{\textbf{Depth-guided Dual Projection View Transformer.}}
    \label{fig:depth}
\vspace{-3mm}
\end{figure}
This section presents our approach for enhancing geometric representations through explicit depth guidance. As illustrated in \cref{fig:arch}, we simultaneously extract depth features $\mathbf{F}^{(D)} = \{\mathbf{F}^D_i \mid i \in [1, N_c]\}$ and depth maps $\{\mathbf{D}_i \mid i \in [1, N_c]\}$ from multi-view images using MoGe-2~\cite{wang2025moge2} while performing 2D image feature extraction in the Image Encoder. Building upon these dense depth cues, we design a depth-guided 2D-to-3D view transformer to reconcile geometric misalignment issues inherent in existing projection paradigms.
However, leveraging depth cues for geometric enhancement is non-trivial: naive depth supervision approaches suffer from domain gaps and scale sensitivity. To identify the optimal integration strategy, we systematically explore multiple approaches and present our successful practice that achieves both geometric accuracy and computational efficiency.

\subsubsection{Incorporating Depth Cues}
Recent depth estimation methods such as MoGe~\cite{wang2025moge,wang2025moge2} achieve strong generalization across diverse scenes. We hypothesize that this depth generalization capability can be directly transferred to occupancy prediction. However, naive approaches that treat depth as explicit 3D signals (RGBD-style) or convert it to pseudo point clouds (LiDAR-style) face fundamental limitations: domain gaps between monocular depth and the occupancy domain, as well as sensitivity to depth scale variations. Notably, directly using MoGe depth for forward projection paradoxically degrades performance, as this leads to image features lacking implicit constraints from depth information, thereby generating low-quality voxel features.  These findings are supported by ablation studies provided in the supplementary material.

We make a critical observation: approximately 90\% of voxels in the occupancy grid are empty, suggesting that directly fitting all voxels is inefficient. Instead, we leverage MoGe's depth information to identify occupied voxels, creating a geometry-aware occupancy mask $M$ that guides the model to focus on non-empty voxels. To generate this mask, we voxelize the depth-derived pseudo point cloud $\mathcal{P}$ through camera projection:
\begin{align}
\mathbf{x}_{\text{cam}}^T &= d \cdot\mathbf{K}_i^{-1} \begin{bmatrix} u & v & 1 \end{bmatrix}^T, \quad d = \mathbf{D}_i(u,v), \label{eq:backprojection} \\
\mathbf{p}_i &= \mathbf{R}_i^\top (\mathbf{x}_{\text{cam}} - \mathbf{t}_i), \label{eq:transform} \\
M(\mathbf{v}) &= \begin{cases} 1, & \text{if } \mathbf{v} \in \text{Voxelize}(\mathcal{P}, r), \\ 0, & \text{otherwise}, \end{cases} \label{eq:occupancy}
\end{align}
where $\mathbf{v}$ denotes voxel center coordinates, $r$ is voxel resolution.

\subsubsection{Depth-guided Dual Projection View Transformer}

Building upon the geometry-aware occupancy mask $M(\mathbf{v})$, we design the \textbf{Depth-guided Dual-Projection View Transformer (D$^2$-VFormer)} to systematically integrate depth cues into dual projection pathways. The core design principle is to leverage the occupancy mask for identifying non-empty voxels, then selectively refine their features through depth-guided densification and semantic enhancement, yielding geometrically accurate and semantically rich voxel representations while avoiding wasteful computation on empty space.
As illustrated in Fig.~\ref{fig:arch}, D$^2$-VFormer comprises three progressive refinement stages:

\noindent\textbf{Stage 1: Forward Projection and Downsampling.} Following BEVStereo~\cite{li2023bevstereo}, we fuse multi-frame image features to lift 2D features into voxel space via depth projection, producing sparse representations with ~30\% occupancy. We downsample both voxel features $\mathbf{F}_{\text{down}}$ and geometry mask $\mathcal{M}_{\text{down}}$ by a factor of $\lambda$ to achieve: (1) \textit{computational efficiency} and (2) \textit{improved depth robustness}—coarser voxel resolution naturally tolerates pixel-level depth estimation errors.

\noindent\textbf{Stage 2: Backward Projection Densification.} As illustrated in Fig.~\ref{fig:depth}, we employ a deformable cross-attention (DCA)~\cite{zhu2020deformable} module to fuse multi-view image features, recovering geometric completeness:
\begin{equation}
\mathbf{F}_{\text{dense}} = \mathrm{DCA}(\mathbf{F}_{\text{down}}, \mathbf{F}^{(I)}),
\end{equation}
However, naive densification without geometric grounding introduces inconsistent features in regions lacking geometric evidence, motivating Stage 3.

\noindent\textbf{Stage 3: Depth-Guided Non-Empty Voxel Refinement.} This stage performs selective two-step refinement on geometrically plausible regions identified by $\mathcal{M}_{\text{down}}$.

\textit{Geometric Refinement.} We enhance geometric consistency by fusing depth features into occupied voxels:
\begin{equation}
\scalebox{0.8}{$\displaystyle
\mathbf{F}_{\text{geo}}(\mathbf{v}) = \begin{cases} \mathrm{DCA}(\mathbf{F}_{\text{dense}}, \mathbf{F}^{(D)}; \mathcal{M}_{\text{down}}), & M_{\text{down}}(\mathbf{v}) = 1, \\ \mathbf{e}_{\text{empty}}, & \text{otherwise}, \end{cases}
$}
\end{equation}
Depth features $\mathbf{F}^{(D)}$ enforce spatial consistency in occupied regions, while learnable empty voxel embeddings $\mathbf{e}_{\text{empty}}$ suppress wasteful learning on empty space.

\textit{Semantic Enhancement.} We refine semantic information by fusing multi-view image features for occupied voxels:
\begin{equation}
\mathbf{F}_{\text{out}} = \mathrm{DCA}(\mathbf{F}_{\text{geo}}, \mathbf{F}^{(I)}; \mathcal{M}_{\text{down}}),
\end{equation}
This progressive two-step refinement concentrates computational resources on semantically meaningful voxels, achieving geometric accuracy and semantic richness without wasteful computation on empty space.

\subsection{Region-guided Semantic Enhancement}
While D$^2$-VFormer enhances geometric representations, semantic prediction in occupancy scenes faces a critical challenge: long-tail category distribution, where rare semantic classes appear far less frequently than common ones, leading to biased semantic learning. Furthermore, we observe that different semantic classes exhibit strong positional preferences in 3D space. This spatial anisotropy of semantics therefore motivates the adaptive allocation of model capacity across regions rather than uniformly. We leverage this observation to design \textbf{Region-guided Expert Transformer (R-EFormer)} and its recursive variant \textbf{R$^2$-EFormer}, which guide region-specific feature refinement based on spatial semantic structure.

\begin{figure*}[t]
    \centering
    \includegraphics[width=\textwidth]{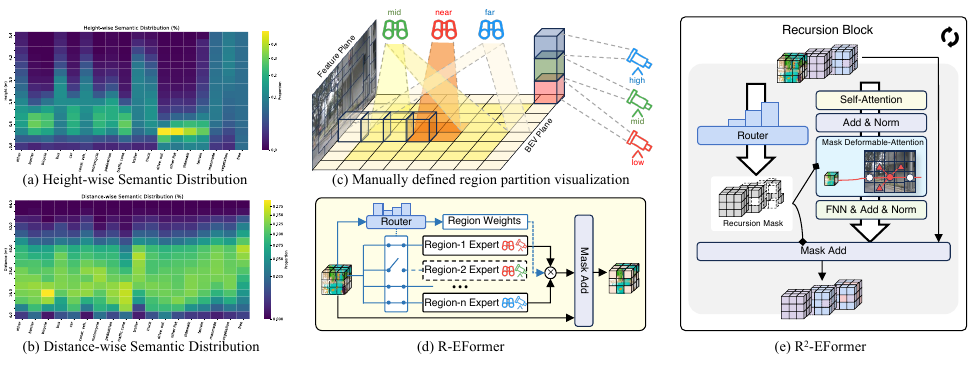}
    \vspace{-2em}
    \caption{\textbf{(a–b)} Spatial semantic distribution reveals strong anisotropy across height and distance. \textbf{(c–d)} R-EFormer partitions 3D space into a 3×3 spatial grid along range (near, mid, far) and height (low, mid, high) dimensions, assigning each region a dedicated expert. \textbf{(e)} R$^2$-EFormer adaptively refines regions through recursive masking.}
    \label{fig:region}
\vspace{-3mm}
\end{figure*}

\subsubsection{Anisotropic Spatial Semantics}
\label{sec:anisotropy}
We first examine the spatial distribution of semantic categories in occupancy space. As shown in \cref{fig:region}~(a–b), semantic categories exhibit strong spatial anisotropy: road surfaces concentrate at low heights and short ranges, vegetation and buildings occupy higher elevations and mid-ranges, while dynamic objects appear only in narrow spatial bands. This spatial anisotropy motivates us to partition the 3D space into regions aligned with semantic distributions. Within each region, the semantic composition becomes more concentrated, effectively reducing the relative imbalance of rare classes and strengthening supervision signals for minority categories. Therefore, applying uniform model capacity across all regions is inefficient; instead, capacity should be allocated region-specifically to match local semantic distributions and enhance learning for rare spatial-semantic combinations.

\subsubsection{Region-guided Expert Transformer}
\label{sec:moe}

Inspired by Mixture-of-Experts (MoE)~\cite{zhou2022moe}, we design R-EFormer, which allocates region-specific experts to capture spatial semantic variations. As shown in \cref{fig:region}~(c–d), the 3D space is partitioned into multiple regions along distance from the ego vehicle and height dimensions, with each region $\mathcal{R}_m$ assigned a dedicated expert $E_m$.
A router network evaluates the importance of each region and computes scores for all regions:
\begin{equation}
s_m = \mathrm{Router}(\mathbf{F}_{\text{out}}), \quad m = 1, \dots, M,
\end{equation}
where $s_m$ is the importance score for region $m$ and $\mathbf{F}_{\text{out}}$ is the input feature from D$^2$-VFormer. The router then selects the top-$K$ most relevant regions:
\begin{equation}
\mathcal{S} = \mathrm{TopK}(\{s_m\}_{m=1}^M, K),
\end{equation}
where $\mathcal{S}$ denotes the set of selected region indices and $K$ controls the number of activated experts. Each expert $E_m$ applies the same DCA module as in D$^2$-VFormer, but restricted to its corresponding region via a binary mask $\mathcal{M}_m$:
\begin{equation}
E_m(\mathbf{F}_{\text{out}}, \mathbf{F}^{(I)}; \mathcal{M}_m) = \mathrm{DCA}(\mathbf{F}_{\text{out}}, \mathbf{F}^{(I)}; \mathcal{M}_m).
\end{equation}
The final output combines weighted predictions from selected experts:
\begin{equation}
\mathbf{F}_{\text{final}} = \sum_{m \in \mathcal{S}} w_m \cdot E_m(\mathbf{F}_{\text{out}}, \mathbf{F}^{(I)}; \mathcal{M}_m),
\end{equation}
where $w_m$ is the normalized weight for region $m$ computed from the importance scores, $E_m$ is the expert dedicated to region $\mathcal{R}_m$, and $\mathcal{M}_m$ is the binary spatial mask defining region $\mathcal{R}_m$. By dedicating experts to predefined regions and selectively activating them based on importance, R-EFormer concentrates model capacity on semantically important regions while handling spatial semantic variations.

\subsubsection{Region-guided Recursive Expert Transformer}
\label{sec:mor}
R-EFormer requires manual region definition, which introduces hyperparameter sensitivity. To enable adaptive region discovery, we further propose R$^2$-EFormer, extending R-EFormer following Mixture-of-Recursions (MoR) principles~\cite{bae2025mor}. Rather than maintaining multiple region-specific experts, R$^2$-EFormer employs a single expert that iteratively refines features over $n$ iterations, with each iteration progressively focusing on refined spatial regions.

Let $\mathbf{F}^{(t)}$ denote the voxel features at iteration $t$ and $\mathcal{M}^{(t)}$ denote the spatial mask in iteration $t$. The mask $\mathcal{M}^{(t)}$ is generated by a router $\mathcal{R}^{(t)}$ that progressively focuses on the most salient regions. Specifically, the router first produces an importance weight map based on the input features and previous mask, from which the top-$k_t$ voxels are selected:
\begin{equation}
\scalebox{0.8}{$\displaystyle
\mathcal{M}^{(t)} = 
\begin{cases}
\Omega, & t = 1, \\
\mathrm{TopK}(\mathcal{R}^{(t)}(\mathbf{F}^{(t-1)}, \mathcal{M}^{(t-1)}), k_t), & t > 1.
\end{cases}
$}
\end{equation}
where $\Omega$ denotes the full voxel space, and $\mathrm{TopK}(\cdot, k_t)$ selects the top-$k_t$ voxels based on these weights. The sequence $k_1 > k_2 > \cdots > k_n$ is a fixed decreasing sequence, ensuring $\mathcal{M}^{(t)} \subset \mathcal{M}^{(t-1)}$.
Given the mask $\mathcal{M}^{(t)}$, the feature refinement at each iteration is performed by a DCA module:
\begin{equation}
\mathbf{F}^{(t)} = \mathrm{DCA}(\mathbf{F}^{(t-1)}, \mathbf{F}^{(I)}; \mathcal{M}^{(t)}), \quad t = 1, \dots, n,
\end{equation}
where $\mathbf{F}^{(t-1)}$ is the feature output from the previous iteration. This iterative process enables the model to concentrate computation on increasingly important voxels while reusing and deepening feature representations from prior iterations.


Compared to R-EFormer, R$^2$-EFormer offers three advantages: it reduces parameters by using a single recursive expert, alleviates sensitivity to manual region definitions through adaptive refinement, and progressively refines semantic predictions by iteratively focusing on high-confidence regions.

\subsection{3D Occupancy Decoder}
The refined features $\mathbf{F}_{\text{final}}$ are upsampled to the target resolution $(X, Y, Z)$ via trilinear interpolation, then passed through a simple CNN-based decoder to produce the final voxel-wise semantic occupancy prediction $\hat{\mathbf{O}}$.

%% file: sec/4_experiments.tex
\section{Experiments}
\subsection{Experiment Setup}
\input{tables/mainresults_MIoU_Occ3d}
\paragraph{Datasets and Evaluation Metrics.}
The nuScenes~\cite{caesar2020nuscenes} dataset contains 1{,}000 driving scenes (700 train, 150 val, 150 test) with approximately 40{,}000 frames, each equipped with a 32\, beam LiDAR and six multi\mbox{-}view RGB cameras. 
Occ3D\mbox{-}nuScenes~\cite{tian2023occ3d} extends nuScenes with dense voxel\mbox{-}wise semantic occupancy annotations but excludes the 150 test scenes due to missing LiDAR, resulting in 850 annotated scenes. 
Annotations in Occ3D\mbox{-}nuScenes are provided for a three\mbox{-}dimensional volume measuring $80\,\mathrm{m} \times 80\,\mathrm{m} \times 6.4\,\mathrm{m}$ in the ego coordinate frame. This spatial range corresponds to $-40\,\mathrm{m}$ to $40\,\mathrm{m}$ along both the X\mbox{-}axis and the Y\mbox{-}axis, and from $-1\,\mathrm{m}$ to $5.4\,\mathrm{m}$ along the Z\mbox{-}axis. The occupancy annotations have a uniform voxel resolution of $0.4\,\mathrm{m}$ on all three axes. Semantic labels span a total of 18 categories, including 16 known object classes, one class representing empty space, and one additional category introduced in the benchmark extension. For evaluation, the benchmark reports the Intersection\mbox{-}over\mbox{-}Union (IoU) metric to assess binary occupancy classification between empty and occupied voxels, which reflects the accuracy of geometric reconstruction. In addition, the mean Intersection\mbox{-}over\mbox{-}Union (mIoU) is computed as the average IoU over all semantic classes, serving as the primary metric for measuring semantic understanding.
\vspace{-0.5em}
\paragraph{Implementation Details.}
Following common practice\cite{huang2022bevdet4d, flashocc, li2023fbocc, ma2024cotr}, we adopt \texttt{ResNet50}~\cite{resnet} as the image encoder and \texttt{moge-2-vits-normal}~\cite{wang2025moge2} for depth estimation. For forward projection, the voxel channel dimension is set to $C=32$ with a spatial resolution of $200 \times 200 \times 16$, covering a physical space of $80 \times 80 \times 6.4$ meters with a voxel size of $0.4$ meters in all dimensions. The projected features are downsampled by a factor of $1/16$, resulting in a voxel grid of size $[H/4, W/4, Z]$.
For the R-EFormer, spatial regions are partitioned by height intervals $\{-1.0 \text{ m} \le h < 0.2 \text{ m},\ 0.2 \text{ m} \le h < 2.2 \text{ m},\ 2.2 \text{ m} \le h \le 5.4 \text{ m}\}$ and distance intervals $\{0 \text{ m} \le d < 10 \text{ m},\ 10 \text{ m} \le d < 30 \text{ m},\ d \ge 30 \text{ m}\}$. For the R$^2$-EFormer, the mask activation ratio is progressively reduced over three iterations, with coverage ratios of $100\%$, $75\%$, and $50\%$ in the first, second, and third iterations, respectively. In the transformer modules, the multi-head attention is configured with $8$ heads and $N_{\text{ref}}=4$ reference points.
All models are trained for $24$ epochs using the AdamW optimizer with a learning rate of $1\times 10^{-4}$, a weight decay of $1\times 10^{-2}$, and a batch size of $2$ per GPU on $8$ NVIDIA L20 GPUs (total batch size of $16$).

\subsection{Comparison with State-of-the-art Methods}
In Table~\ref{table:main_results_miou_on_occ3d}, we present quantitative comparisons on the Occ3D-nuScenes~\cite{tian2023occ3d} benchmark against existing open-source state-of-the-art methods across three representative paradigms: forward-projection methods (FlashOcc~\cite{flashocc}, BEVDet4D~\cite{huang2022bevdet4d}), backward-projection frameworks (BEVFormer~\cite{li2022bevformer}, CTF-Occ~\cite{tian2023occ3d}, TPVFormer~\cite{huang2023tri}, OSP~\cite{shi2024occupancysetpoints}, ViewFormer~\cite{li2024viewformer}), and dual-projection designs (FB-Occ~\cite{li2023fbocc}, COTR~\cite{ma2024cotr}).
Our Dr.Occ framework, built upon BEVDet4D~\cite{huang2022bevdet4d} with depth-guided dual projection and region-guided expert transformer, achieves the highest overall mIoU with substantial gains of +7.43\% mIoU and +3.09\% IoU, surpassing all publicly available approaches.
Our method achieves significant improvement on foreground IoU, along with a steady gain on background classes, demonstrating effective mitigation of class imbalance.
Beyond our complete framework, we validate the plug-and-play capability by integrating our modules into COTR~\cite{ma2024cotr} (reproduced under identical conditions as official weights are unavailable), achieving an additional 1\% mIoU improvement.
This consistent enhancement across diverse baseline methods substantiates the effectiveness and versatility of our depth-guided dual projection coupled with region-guided expert transformer.

\subsection{Ablation Study}
\input{tables/ablation_components}
To evaluate the contribution of each proposed module, we perform an ablation study on the Occ3D-nuScenes~\cite{tian2023occ3d} dataset, using BEVDet4D~\cite{huang2022bevdet4d} as the baseline framework.
\vspace{-2em}
\paragraph{The Effectiveness of Each Component.}
The ablation results on Occ3D-nuScenes are summarized in Table~\ref{tab:ablation_compo}. 
The baseline model achieves 70.36\% IoU and 36.01\% mIoU. 
Introducing the geometric enhancement module D\textsuperscript{2}-VFormer yields clear gains of +0.93\% IoU and +5.44\% mIoU, verifying its effectiveness in geometric completion and semantic improvement. 
Combining D\textsuperscript{2}-VFormer with the semantic decoder R-EFormer further raises performance to 73.45\% IoU and 43.03\% mIoU, highlighting the complementarity between geometric and semantic enhancement. 
Replacing R-EFormer with its recursive variant R\textsuperscript{2}-EFormer slightly reduces IoU but achieves the highest mIoU (43.43\%), owing to its stronger focus on rare and hard-to-recognize categories through adaptive recursive refinement.
\begin{figure}[t]
    \centering
    \includegraphics[width=\linewidth]{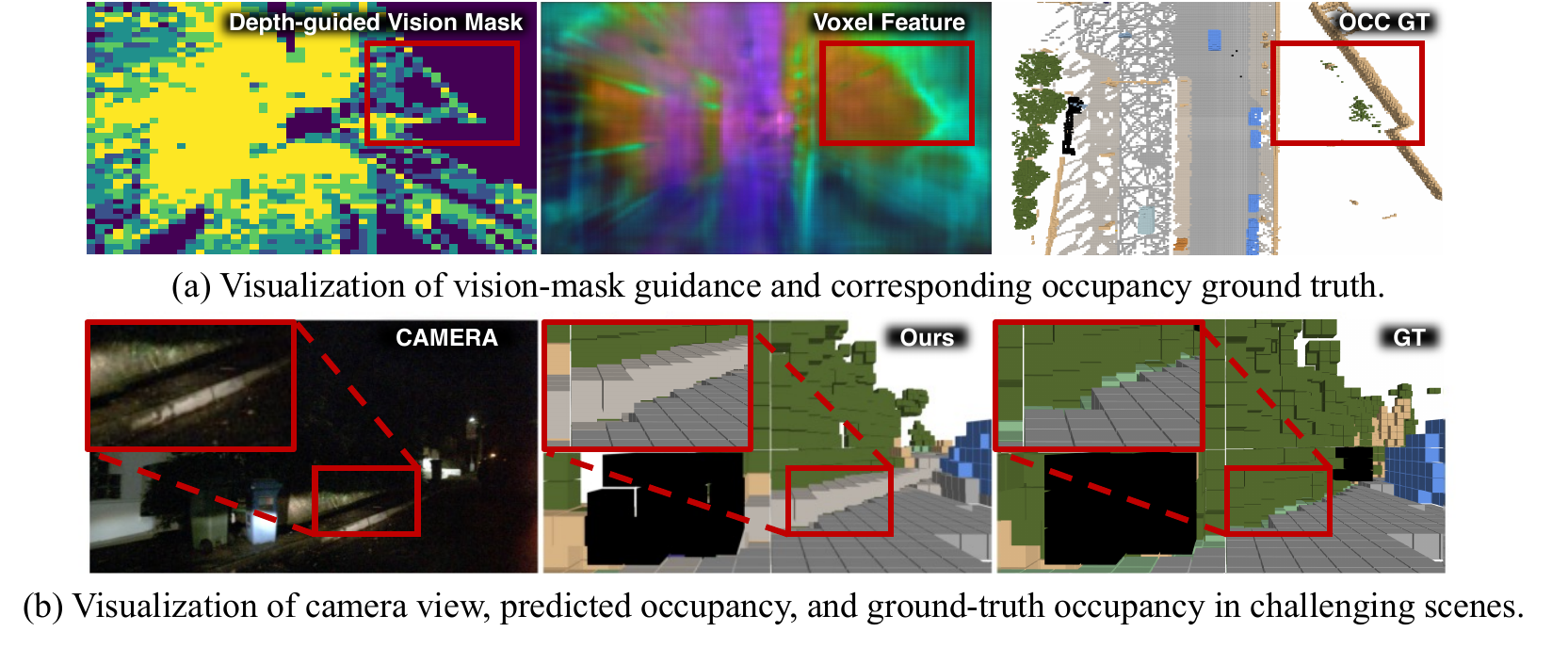}
    \vspace{-1em}
    \caption{\textbf{Benefits of depth-guided dual projection.} Our D\textsuperscript{2}-VFormer generates more complete and detailed occupancy predictions (b) by learning geometry-aware features (a), as evidenced by the closer alignment with ground truth.}
    \label{fig:ab-dv}
\vspace{-1em}
\end{figure}
\vspace{-0.5em}
\paragraph{The Effectiveness of D$^2$-VFormer.}
To assess the effectiveness of the depth-guided dual projection in capturing geometric structures, \cref{fig:ab-dv} presents qualitative results. 
Panel~(a) illustrates the occupancy mask, corresponding feature maps after applying our module, and occupancy ground truth, showing that depth-aware projections align feature responses with scene geometry. 
Panel~(b) compares camera inputs, predicted occupancy, and ground truth, where introducing D\textsuperscript{2}-VFormer enables the model to reconstruct more complete drivable areas and recover fine details such as pedestrian walkways. 
These visual observations align with the quantitative improvements in Table~\ref{tab:ablation_compo}, confirming that depth-guided dual projection strengthens both geometric completion and semantic estimation.

\begin{figure}[t]
    \centering
    \includegraphics[width=\linewidth]{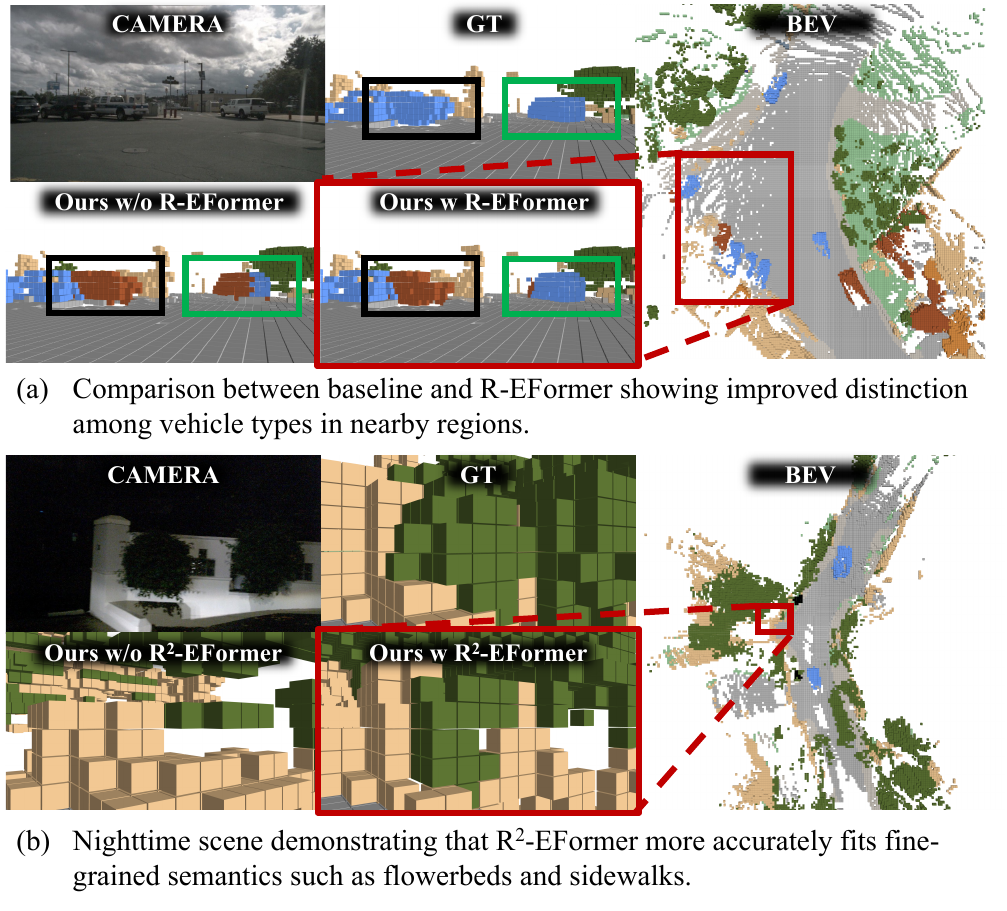}
    \caption{\textbf{Qualitative visualization of region-guided semantic enhancement.} Our method produces more accurate and human-interpretable occupancy predictions by adaptively partitioning regions and allocating specialized experts.}
    \label{fig:ab-re}
\vspace{-1.5em}
\end{figure}

\vspace{-0.5em}
\paragraph{The Effectiveness of R-EFormer.}
To evaluate the impact of region-guided expert modeling, we visualize qualitative comparisons in \cref{fig:ab-re}~(a). As shown, R-EFormer achieves notable improvements: for the center vehicle, although the ground truth annotation is incorrect, R-EFormer not only correctly identifies it but also captures sharper semantic boundaries and finer trunk details compared to the baseline; for the right vehicle, R-EFormer delivers more complete and accurate predictions than the baseline. These improvements demonstrate that region-specific experts effectively enhance both semantic precision and geometric completeness in densely clustered areas.

\vspace{-0.5em}
\paragraph{The Effectiveness of R$^2$-EFormer.}
R$^2$-EFormer enhances semantic prediction by iteratively refining attention from the full voxel space to progressively smaller masks, 
thereby concentrating on ambiguous or hard-to-distinguish regions.  
As illustrated in \cref{fig:ab-re}~(b), in a challenging night scene with complex background structures such as flowerbeds and sidewalks, R$^2$-EFormer adaptively focuses attention on hard-to-fit fine-grained voxels, producing smoother reconstructions and more accurate semantics. 
This adaptive recursion further strengthens the modeling of small and ambiguous regions, leading to coherent and reliable occupancy predictions beyond fixed-region decoding.

%% file: tables/mainresults_MIoU_Occ3d.tex
\begin{table*}[t]
  \small
  \setlength{\tabcolsep}{0.0030\linewidth}
  \centering
  \begin{tabular}{l | cc  | >{\columncolor{Gray}}c | c c c c c c c c c c c c c c c c c c}
      \toprule
      Method
      & \rotatebox{90}{Backbone}
      & \rotatebox{90}{Input Size}
      & \rotatebox{90}{mIoU($\%$)}
      & \rotatebox{90}{\textcolor{otherscolor}{$\blacksquare$} others} 
      & \rotatebox{90}{\textcolor{barriercolor}{$\blacksquare$} barrier} %
      & \rotatebox{90}{\textcolor{bicyclecolor}{$\blacksquare$} bicycle} %
      & \rotatebox{90}{\textcolor{buscolor}{$\blacksquare$} bus} %
      & \rotatebox{90}{\textcolor{carcolor}{$\blacksquare$} car} %
      & \rotatebox{90}{\textcolor{constructcolor}{$\blacksquare$} cons. veh.} %
      & \rotatebox{90}{\textcolor{motorcolor}{$\blacksquare$} motor.} %
      & \rotatebox{90}{\textcolor{pedestriancolor}{$\blacksquare$} pedes.} %
      & \rotatebox{90}{\textcolor{trafficcolor}{$\blacksquare$} tfc. cone} %
      & \rotatebox{90}{\textcolor{trailercolor}{$\blacksquare$} trailer} %
      & \rotatebox{90}{\textcolor{truckcolor}{$\blacksquare$} truck} %
      & \rotatebox{90}{\textcolor{driveablecolor}{$\blacksquare$} drv. surf.} %
      & \rotatebox{90}{\textcolor{otherflatcolor}{$\blacksquare$} other flat} %
      & \rotatebox{90}{\textcolor{sidewalkcolor}{$\blacksquare$} sidewalk} %
      & \rotatebox{90}{\textcolor{terraincolor}{$\blacksquare$} terrain} %
      & \rotatebox{90}{\textcolor{manmadecolor}{$\blacksquare$} manmade} %
      & \rotatebox{90}{\textcolor{vegetationcolor}{$\blacksquare$} vegetation} \\ %
      \midrule
      BEVFormer~\cite{li2022bevformer}              & R101  & 1600$\times$900   & 26.9 & 5.9 & 37.8 & 17.9 & 40.4 & 42.4 & 7.4 & 23.9 & 21.8 & 21.0 & 22.4 & 30.7 & {55.4} & 28.4 & 36.0 & 28.1 & 20.0 & 17.7 \\
      CTF-Occ~\cite{tian2023occ3d}                  & R101  & 1600$\times$900    & 28.5 & 8.1 & 39.3 & 20.6 & 38.3 & 42.2 & 16.9 & 24.5 & 22.7 & 21.1 & 23.0 & 31.1 & 53.3 & 33.8 & 38.0 & 33.2 & 20.8 & 18.0 \\
      TPVFormer~\cite{huang2023tri}            & R101  & 1600$\times$900   & 27.8 & 7.2 & 38.9 & 13.7 & 40.8 & 45.9 & 17.2 & 20.0 & 18.9 & 14.3 & 26.7 & 34.2 & 55.7 & 35.5 & 37.6 & 30.7 & 19.4 & 16.8 \\
      SparseOcc~\cite{liu2023fully}       & R50   & 704$\times$256   & 30.9 & 10.6 & 39.2 & 20.2 & 32.9 & 43.3 & 19.4 & 23.8 & 23.4 & 29.3 & 21.4 & 29.3 & 67.7 & 36.3 & 44.6 & 40.9 & 22.0 & 21.9 \\
      \hline
      OSP$^{\ast}$~\cite{shi2024occupancysetpoints}               & R101  & 1600$\times$900    & 41.2 & 10.9 & 49.0 & 27.7 & 50.2 & 55.9 & 22.9 & 31.0 & 30.9 & 30.3 & 35.6 & 31.2 & 82.1 & 42.6 & 51.9 & 55.1 & 44.8 & 38.2 \\
      FlashOcc$^{\ast}$\cite{flashocc} & R50 & 704$\times$256  &
      37.8 & 9.1 & 46.3 & 17.7 & 42.7 & 50.6 & 23.7 & 20.1 & 22.3 
      & 24.1 & 30.3 & 37.4 & 81.7 & 40.1 & 52.3 & 56.5 & 47.7 & 40.6 \\
      FB-Occ$^{\ast}$~\cite{li2023fbocc}         & R50   & 704$\times$256    & 39.1 & 13.6 & 44.7 & 27.0 & 45.4 & 49.1 & 25.2 & 26.3 & 27.9 & 27.8 & 32.3 & 36.8 & 80.1 & 42.8 & 51.2 & 55.1 & 42.2 & 37.5 \\
      ViewFormer$^{\ast}$~\cite{li2024viewformer} & R50   & 704$\times$256     & 41.9 & 12.9 & 50.1 & 27.9 & 44.6 & 52.9 & 22.4 & 29.6 & 28.0 & 29.2 & 35.2 & 39.4 & \textbf{84.7} & \textbf{49.4} & \textbf{57.4} & \textbf{59.7} & 47.4 & 40.6 \\
      \hline
      BEVDet4D$^{\ast}$\cite{huang2022bevdet4d} & R50 & 704$\times$256  &
      36.0 & 8.2  & 44.2 & 10.3 & 42.1 & 49.6 & 23.4 & 17.4 & 21.5
      & 19.7 & 31.3 & 37.1 & 80.1 & 37.4 & 50.4 & 54.3 & 45.6 & 39.6 &  \\
      BEVDet4D(+Dr.Occ)$^{\ast}$ & R50 & 704$\times$256  & 43.4 & \textbf{14.9} & 52.8 & 30.7 & 45.6 & \textbf{55.1} &  30.3 & 30.8 & \textbf{31.0} & \textbf{35.9} & \textbf{39.1} & \textbf{42.5} & 81.1 & 43.4 & 54.0 & 57.6 & 49.0 & 44.5 \\
      \hline
      COTR$^{\ast}$$^{\ddagger}$~\cite{ma2024cotr}   & R50   & 704$\times$256    & 43.1 & 12.8 & 51.6 & 30.3 & 45.4 & 54.9 & \textbf{30.4} & 30.4 & 30.0 & 33.9 & 36.9 & 42.1 & 83.4 & 44.4 & 55.3 & 58.0 & 49.6 & 44.0 \\
      COTR(+Dr.Occ)$^{\ast}$ & R50 & 704$\times$256  & \textbf{44.1} & 14.0 & \textbf{53.0} & \textbf{32.6} & \textbf{45.8} & 54.9 & 29.7 & \textbf{31.5} & 30.9 & 35.7 & 38.0 & 42.5 & 83.8 & 44.9 & 56.1 & 58.8 & \textbf{51.1} & \textbf{45.6} \\

  \bottomrule
  \end{tabular}
  \caption{\textbf{Quantitative comparison of mIoU (\%) on Occ3D-nuScenes\cite{tian2023occ3d}.} $^{\ast}$ indicates models trained with camera mask supervision. $^{\ddagger}$ indicates methods re-trained in this work as official pre-trained weights are not available.}

  \label{table:main_results_miou_on_occ3d}
\vspace{-1.5em}
\end{table*}

%% file: tables/ablation_components.tex
\begin{table}[tp]
    \resizebox{\linewidth}{!}{

        \begin{tabular}{c|cc|cc}
        \toprule
        \multicolumn{1}{c|}{Geometric} & \multicolumn{2}{c|}{Semantic} & \multicolumn{2}{c}{Metric} \\
        \midrule
        
        D$^2$-VFormer & R-EFormer & R$^2$-EFormer & IoU($\%$)$\uparrow$ & mIoU($\%$)$\uparrow$ \\
        \midrule
                  &            &                      & 70.36   & 36.01 \\
        \ding{51} &            &                     & 71.29  & 41.45  \\
        \ding{51} &  \ding{51} &                    & \textbf{73.45}   & 43.03   \\
        \rowcolor[HTML]{EFEFEF}
        \ding{51} &  & \ding{51}   & 72.87   & \textbf{43.43}     \\
    
        \bottomrule
        \end{tabular}
    }
    \caption{\textbf{Ablation study on the each component.}}
    \label{tab:ablation_compo}
    \vspace{-1em}
\end{table}

%% file: sec/5_conclusion.tex
\section{Conclusion}
In this paper, we present Dr.Occ, a unified vision-based 3D occupancy prediction framework that jointly addresses geometric reconstruction and semantic imbalance through depth-guided feature mapping and region-guided expert modeling. By leveraging advanced depth priors and spatial anisotropy of semantic distributions, Dr.Occ enables dense, geometrically consistent, and semantically balanced occupancy estimation. Extensive experiments on the Occ3D-nuScenes benchmark demonstrate that Dr.Occ significantly improves various baseline models, achieving 7.43\% mIoU gains on BEVDet4D and further boosting state-of-the-art COTR by 1.0\% mIoU. We believe that Dr.Occ provides a new perspective for joint geometric-semantic modeling and will inspire future exploration in vision-based 3D perception for autonomous driving.

\section*{Acknowledgement}
The work was supported by NSFC (62301370), GuangDong Basic and Applied Basic Research Foundation (2025A1515010440), and Shenzhen Science and Technology Program (JCYJ20240813111202004).